\documentclass[preprint,review,12pt]{elsarticle}

\usepackage{hyperref}
\usepackage{booktabs}
\usepackage{multirow}
\graphicspath{{imgs/}} 
\usepackage{overpic}
\usepackage{subcaption}
\usepackage{array}

\usepackage{amsfonts} 
\usepackage{amssymb}
\usepackage{breqn}

\usepackage{lineno}

\usepackage{color} 
\newcommand\chg[1]{{\color{black}#1}}

\usepackage{array}
\usepackage{multicol}
\usepackage{makecell}

\usepackage{soul}
\usepackage{physics}
\usepackage{cleveref}

\journal{Pattern Recognition}
\begin{document}
\begin{frontmatter}


\title{GNN-LoFI: a Novel Graph Neural Network through Localized Feature-based Histogram Intersection}

\author[unive]{Alessandro Bicciato}
\author[unive,usi]{Luca Cosmo}
\author[unive]{Giorgia Minello}
\author[polyu]{Luca Rossi\corref{cor1}}
\author[unive]{Andrea Torsello}
\affiliation[unive]{organization={Ca' Foscari University of Venice},country={Italy}}
\affiliation[usi]{organization={University of Lugano},country={Switzerland}}
\affiliation[polyu]{organization={The Hong Kong Polytechnic University},country={Hong Kong}}

\cortext[cor1]{Corresponding author: luca.rossi@polyu.edu.hk}
\begin{abstract}
Graph neural networks are increasingly becoming the framework of choice for graph-based machine learning. In this paper, we propose a new graph neural network architecture that substitutes classical message passing with an analysis of the local distribution of node features. To this end, we extract the distribution of features in the egonet for each local neighbourhood and compare them against a set of learned label distributions by taking the histogram intersection kernel. The similarity information is then propagated to other nodes in the network, effectively creating a message passing-like mechanism where the message is determined by the ensemble of the features. We perform an ablation study to evaluate the network's performance under different choices of its hyper-parameters. Finally, we test our model on standard graph classification and regression benchmarks, and we find that it outperforms widely used alternative approaches, including both graph kernels and graph neural networks.
\end{abstract}




\end{frontmatter}

\section{Introduction}
 
The beginning of the 21st century has undoubtedly seen deep learning, a re-branding of artificial neural networks, made possible by a series of pivotal discoveries in the late '90s and early '00s~\cite{hochreiter1997long,hinton2007learning,bengio2012practical}, taking the center stage in the field of machine learning. Significant advances in terms of available hardware were key to this sudden explosion in popularity, with GPUs speeding up training algorithms heavily reliant on matrix multiplications by orders of magnitude~\cite{oh2004gpu}. As a result, models based on different flavours of the Transformer architecture~\cite{vaswani2017attention,devlin2019BERT} and Convolutional Neural Networks (CNNs)~\cite{guo2022cmt} are now the de-facto standards to solve problems in fields that range from computer vision to natural language processing. However seemingly unstoppable these models may appear to be, they still encounter a number of difficulties when dealing with data that is characterized by a rich structure. A recent example is that of latent diffusion models, whose ability to grasp even basic relations between agents and objects has been put in doubt by recent studies~\cite{conwell2022testing}.

One way to capture this rich structural information is to use graph-based data representations. Graphs are a natural and powerful way to abstract data where structure plays an important role, from images~\cite{johnson2015image} to biological data~\cite{borgwardt2005protein} and collaboration networks~\cite{lima2014coding}. Standard machine learning techniques and even modern deep learning architectures struggle to cope with the rich expressiveness of graph data and thus require substantial modifications to work on this type of data. Over the past few years, Graph Neural Networks (GNNs) have emerged as efficient and effective models to learn on graph data~\cite{scarselli2008graph,atwood2016diffusion,kipf2016semi,gilmer2017neural,velickovic2018graph,bai2020learning,cosmo2021graph,cosmo2020latent,kazi2022differentiable,bicciato2022classifying}, often outperforming traditional approaches based on graph kernels~\cite{shervashidze2011weisfeiler}. The common idea of these architectures is to use the graph structure to guide the spread of the nodes feature information between neighbouring nodes, also knowns as message passing. Compared to graph kernels, the state-of-the-art approach to learning on graphs prior to the advent of modern deep learning architectures, GNNs provide an end-to-end learning framework where input features do not need to be handcrafted anymore but can be learned directly by the model itself given the task at hand. More recently, some works~\cite{gilmer2017neural,xu2018powerful,morris2019weisfeiler} focused on tackling the limited expressive power of 1-hop message passing based architectures, which can be shown to be at most as powerful as standard graph kernels based on the one-dimensional Weisfeiler-Lehman (WL) isomorphism test~\cite{shervashidze2011weisfeiler}.

In this paper, we propose GNN-LoFI, a novel GNN architecture based on a localized feature-based histogram intersection. The fundamental idea underpinning our model is to substitute classical message passing with an analysis of the local distribution of node features.
To this end, we extract the distribution of features in the egonet (i.e. local neighbourhood) for each node and compare them against a set of learned label\footnote{In the remainder of the paper we use the terms \textit{categorical node features} and \textit{labels} interchangeably.} distributions by taking the histogram intersection kernel. The similarity information is then propagated to other nodes in the network, effectively creating a message passing-like mechanism where the message is determined by the ensemble of the features, as opposed to the less sophisticated generation and aggregation functions of standard message passing architectures. Note that by looking at feature distributions over neighbourhoods, the network is also able to capture the local topology of the graph. Moreover, an advantage of this formulation is that the similarity score between the feature distributions over egonets and the set of learned histograms allows us to provide some degree of interpretation of the model decisions, as discussed in Section~\ref{sec:experiments}.

As part of the experimental evaluation, we perform an extensive ablation study to evaluate the impact of the choice of the network hyper-parameters on its performance. We observe that there seems to exist some trade-off between the size of the egonets and the depth of the network, but we otherwise observe that the choice of the optimal hyper-parameters tends to depend on the dataset. Finally, we test our model on both graph classification and regression benchmarks against widely used alternative methods, including both graph kernels and GNNs, verifying that the performance of our model is superior to that of competing ones.

In summary, our main contributions are:
\begin{itemize}
\item We introduce GNN-LoFI, a novel message passing-like model where we define a convolution operation between local neighbourhoods and learned masks in terms of a histogram intersection kernel;
\item The added non-linearities together with the learned masks allow GNN-LoFI to capture more complex and non-linearly separable properties of the neighbourhood;
\item The resulting model maintains a complexity that is comparable to that of simpler message passing models like GCN;
\item Finally, learning the masks allows one to interpret the output of GNN-LoFI, in contrast to standard message passing models that involve non-trivial non-linear transformations of the node features. 
\end{itemize}

\chg{It should be noted that our model is designed for graph-level rather than node-level classification tasks. In node-level classification tasks, where high-dimensional continuous-valued node features play an important role, the use of histograms introduces boundary effects due to the inability to capture the correlation between adjacent discretization buckets. This shortcoming can be addressed using alternative approaches based on the Earth mover’s distance, something we intend to explore in future work.}

The remainder of this paper is organised as follows. Section~\ref{sec:related} provides a summary of related works, while Section~\ref{sec:model} introduces the proposed model. Section~\ref{sec:experiments} presents the results of the ablation study and analyzes its performance on both graph classification and regression benchmarks. Finally Section~\ref{sec:conclusion} concludes the paper and highlights potential directions for future work.

\section{Related work}\label{sec:related}
While graphs provide an information-rich representation for data where structure plays an important role, this comes at the cost of an increased difficulty in applying machine learning techniques that expect a vectorial input. This, in turn, is due to a lack of canonical ordering of the graph nodes, which can vary in number and require permutation-invariant operations or an alignment to a reference structure in order to embed the graphs into a vectorial space. Traditionally, this issue was addressed using kernel methods, where a pairwise positive definite similarity measure is used in lieu of an explicit vectorial representation of the input graph. This can be achieved by finding some set of (typically small) substructures from the two graphs in question and then
quantifying how similar the sets generated from each graph are~\cite{kriege2020survey}. Following this framework, a wide variety of graph kernels has been introduced, which differ mainly in the type of substructures used: from histograms of node labels~\cite{kriege2020survey} to random walks~\cite{zhang2018retgk}, shortest-paths~\cite{borgwardt2005shortest}, or subtrees~\cite{shervashidze2011weisfeiler}.

With the advent of deep learning, the attention of researchers swiftly moved from graph kernels to GNNs. Most of GNNs are based on the idea of propagating node information between groups of nodes. Several rounds of propagation (\textit{i.e.}, layers) produce new node features that capture local structural information and can be used, followed by pooling operations and fully connected layers, for graph classification, node classification, and link prediction.

Scarselli et al.~\cite{scarselli2008graph} are often credited with being the first to introduce this type of neural architecture for graphs by introducing a model where a diffusion mechanism is used to learn the node latent representations in order to exchange neighbourhood information. However, over the past years, a large number of GNNs has been introduced, from convolutional~\cite{kipf2016semi,atwood2016diffusion} to attentional~\cite{velickovic2018graph} ones, with the majority of them falling under the so-called message passing category~\cite{gilmer2017neural}. Despite its popularity, the latter category has been shown to yield models whose expressive power is formally equivalent to that of the WL graph isomorphism test~\cite{xu2018powerful,morris2019weisfeiler}. More recently, a number of researchers have looked into more expressive GNNs that go beyond simple message passing. Bevilaqua et al.~\cite{bevilacqua2022equivariant}, in a way reminiscent of graph kernels, represent graphs as bags of substructures and show that this allows the WL test to distinguish between otherwise indistinguishable graphs. Bouritsas et al.~\cite{bouritsas2022improving} propose a method that belongs to a wider group of approaches using a positional encoding of the graph nodes to improve the ability of the WL test to discriminate between non-isomorphic graphs. Bicciato et al.~\cite{bicciato2022classifying} proposed a message passing-like network where each node of the graph individually learns the optimal message to pass to its neighbours. A few recent works~\cite{nikolentzos2020random,cosmo2021graph,feng2022kergnns} have also started looking at the intersection between graph kernels and GNNs, in an effort to overcome the limitations of message passing architectures in terms of expressive power and interpretability. \chg{Some authors have also started to look into the possibility of using GNNs to cluster both graph~\cite{muller2023graph} and non-graph data~\cite{zhang2022non,muller2023graph}.} For a comprehensive (yet, given the pace the field is developing at, already slightly outdated) survey of GNNs we refer the reader to~\cite{wu2020comprehensive}.

\chg{In Section~\ref{sec:model} we discuss the relation between our network, GNN-LoFI, and the family of message passing networks. Our network is also reminiscent of graph kernels based on histograms of node labels, with the crucial difference that in our model the histograms are learned end-to-end and computed on local node neighbourhoods, rather than simply reflecting the global distribution of features across the graph nodes.}

\section{Method}\label{sec:model}
In this section, we introduce the proposed architecture and its key component, the LoFI layer, in the broader context of GNN architectures. To help the reader, the adopted notation is also displayed in \Cref{tab:notation}.

\begin{table}
\caption{Adopted notation. \textsc{(in)} means it applies only to layer input whereas \textsc{(out)} only to layer output.}
\centering
\renewcommand{\arraystretch}{0.8}
\begin{tabular}{w{c}{1cm} W{l}{6.5cm}  w{l}{1.5cm}}
\toprule

\multicolumn{1}{c}{\textbf{Symbol}} 
& \multicolumn{1}{l}{\textbf{Definition}}
& \multicolumn{1}{l}{\textbf{Dimension}}\\\\

\cmidrule(){1-1}\cmidrule(){2-2}\cmidrule(){3-3} 


$|\mathcal{S}|$            & Cardinality of the set  $\mathcal{S}$   &    \\

$\mathcal{G}$     & Graph & \\

$\mathcal{V}$     & Vertex set   &  $|\mathcal{V}|\; = n$\\

$v$     & Single node $\in \mathcal{V}$  &        \\

$\mathcal{E}$     & Edge set   &  $|\mathcal{E}|\;  $\\

$X$  & Node feature set associated to  $\mathcal{G}$&    $n \times d$\\

$d$     & Layer input feature dimension    &    \\

$L$  & Total number of layers  &     \\

$l$            & Layer, $l=1,\dots, L$  &    \\

$M$ & \makecell[tl]{\textsc{(out)} Layer total number of masks \\and out feature dimension } &   \\

$\mathcal{N}_{G}^r (v)$  &\makecell[tl]{Egonet of radius $r$ centred on node\\  $v$ of input graph $\mathcal{G}$, \textsc{abbr.} $  \mathcal{N}_v$ }    &  $|\mathcal{N}_{G}^r (v)|  $ \\

$X_v$     &  \textsc{(in)} Node feature set associated to 
$  \mathcal{N}_v$   &  $|\mathcal{N}_v| \times d$ \\

$\vectorbold{x}_u$& \textsc{(in)} Node $u$  feature vector &   $1 \times d$ \\

$\mathbf{z}_u $     &  \textsc{(out)} Node $u$ feature vector    &   $1 \times M$ \\
$\mathbf{z}_G $     &  \textsc{(out)} $\mathcal{G}$ vectorial representation    &   $1 \times M$ \\

$D_j$ & \makecell[tl]{$j$-th learned dictionary in a layer, \\with $j=1,\dots,M$ }    &        $W \times d$\\

$W$            & \makecell[tl]{Total number of words (tokens) in the \\dictionary $D_j$}  &    \\

$w_i$            & Word (token)  $\in D_j$, with $i=1,\dots,W$  &   $1 \times d$\\

$\vectorbold{f}_j $ &  \makecell[tl]{$j$-th learned histogram of the layer, \\with $j=1,\dots,M$ }      &      $1 \times W $  \\

$\mathcal{M}_j $ & \makecell[tl]{ $j$-th mask of the layer, defined as  \\$\mathcal{M}_j = (D_j,\vectorbold{f}_j)$,      with $j=1,\dots,M$  } &   \\

$\vectorbold{h}_{v,j}  $   & \makecell[tl]{Feature soft histogram of $X_v$, \\defined as $h(X_v ;D_j )$}  &     $1 \times W$ \\

\bottomrule
\end{tabular}
\label{tab:notation}
\end{table}

\subsection{Learning on graphs}
Given an undirected graph $\mathcal{G}=(\mathcal{V},X,\mathcal{E})$, with $n=|\mathcal{V}|$ nodes, where each node $u\in \mathcal{V} $ is associated to a $d$-dimensional input feature vector $\vectorbold{x}_u$ stored row-wise in the feature matrix $X \in \mathbb{R}^{n \times d}$, and an edge set $\mathcal{E}$, a common strategy in graph classification learning algorithms is to produce a vectorial representation of $\mathcal{G}$ that takes into account both the node features $X$ and the structural information provided by the edge set $\mathcal{E}$. 
Generalizing the convolution operator to graphs, GNNs learn 
a function that embeds the graph into a vector by performing message passing~\cite{gilmer2017neural}. For each node $v \in \mathcal{V}$, a new feature vector $\vectorbold{z}_v$ is derived as
 
\begin{equation}\label{eq:convolution}
\mathbf {z} _{v}= f\left(\mathbf {x}_v,\bigoplus _{u\in N_G^1(v)} g(\mathbf {x}_v,\mathbf x_u)\right)
\end{equation}

where $\mathcal{N}_{G}^1 (v)$ is the egonet of radius 1 centred on the node $v$ of the input graph $\mathcal{G}$ (\textit{i.e.}, the 1-hop neighbourhood of $v$), $f$ and $g$ are differentiable functions with learnable parameters, and $\bigoplus$ is a permutation invariant aggregation operator. $f$ and $g$ are often referred to as the \textit{update} and \textit{message} functions, respectively. The convolution operation of Eq.~\ref{eq:convolution} is repeated for each layer, where the input data for the layer $l$ is given by the output of the previous layer,  $l-1$, \textit{i.e}, $\vectorbold{x}_v^{l} = \vectorbold{z}_v^{l-1} $, for all the $L$ layers except the first one. The final graph-level vectorial representation $\mathbf{z}_G$ is obtained by applying a node-wise permutation invariant aggregation operator on the node features $\mathbf{z}^L_v$, with $v = 1,\dots, n$.

Most message passing methods involve aggregating the features of neighbouring nodes by summing their feature vectors and applying some non-linear transformation. More advanced techniques can also consider edge features in the non-linear transformation, but the aggregation mechanism remains the same. With our model, we propose a different strategy to perform convolution on graphs.

\begin{figure}[t!]
\centering
\includegraphics[trim={0 0cm 0 0},clip,width = 0.9\textwidth]{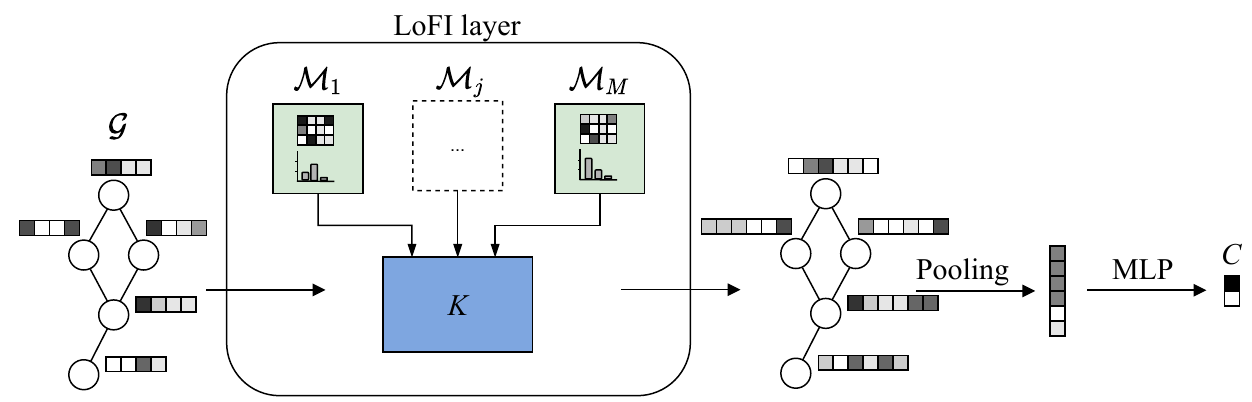}
\caption{The proposed GNN-LoFI architecture. The input graph is fed into one or more LoFI layers, where the feature distributions on egonets centered at each node are compared to a set of learned histograms. The output is a new set of real-valued feature vectors associated with the graph nodes. We obtain a graph-level feature vector through pooling on the nodes features, which is then fed to an MLP to output the final classification label.}
\label{fig:architecture}
\end{figure}

\subsection{The proposed GNN-LoFI architecture}
Unlike the standard aggregation process performed by message passing techniques, we design the convolution operation in terms of a \textit{feature-based histogram intersection} kernel.  Essentially, we compute the similarity between the distribution of features over the neighbourhood of each node and a set of learned feature histograms, in relation  to a set of learned dictionaries. 
The local neighbourhood feature distributions, the learned feature histograms, and the learned dictionaries are the 3 key ingredients of the GNN-LoFI architecture.

Specifically, for each node $v$ of the input graph $\mathcal{G}$ we consider the neighbourhood given by the egonet $\mathcal{N}_G^r(v)$ consisting of the central node $v$ and the nodes at a distance at most $r$ from $v$. We ease the notation by dropping both $r$ and $G$ from the above definition and simply referring to the set of nodes belonging to the egonet centered in $v$ as $\mathcal{N}_v$.

At each layer of the network, we compute, for each egonet, the corresponding node feature distribution with respect to a dictionary of features that we allow our model to learn. 

We can then compute the similarity between the feature distribution and a reference histogram of features, where the latter, like the corresponding dictionary, is learned. Each layer in our model has $M$ such pairs of dictionaries and histograms, resulting in positive real-valued vector 
$\mathbf{z}_v\in \mathbb{R}^
{+M}$ of similarities that define the output of our convolutional operation.

As in standard GNNs architectures, we repeat the convolution operation for a total of $L$ layers and use max-pooling as an order-invariant operation to produce a global descriptor of the graph, which is then given as input to a final MLP layer to obtain the final task-dependent output. Figure~\ref{fig:architecture} shows an overview of the GNN-LoFI architecture, while in the following subsection we discuss more in detail its key components and its relation to standard message passing architectures.

\subsection{Feature-based histogram intersection and the LoFI layer}

For each $v \in \mathcal{V}$, let $X_v = \lbrace \vectorbold{x}_u \mid u \in \mathcal{N}_v \rbrace$ denote the set of input features for the LoFI layer associated to the nodes of the egonet centered in $v$.  As discussed in the previous subsection, our convolution operation is based on computing the similarity between the distribution of node features in $\mathcal{N}_v$ and a set of learned histograms. 

To do so, each LoFI layer needs to learn $M$ pairs of dictionaries and corresponding histograms, \textit{i.e.}, $\mathcal{M}_j = (D_j,\vectorbold{f}_j)$, $j=1, \dots, M$. We call each of these pairs a \textit{mask}.
Each dictionary $D_j$ can be seen as a collection of unique tokens, or \textit{words}, $w_i$ with $i = 1, \dots, W$, represented by vectors lying on the same $d$-dimensional space of the input feature vectors $x_u$. The histogram $\vectorbold{f}_j$ is defined with respect to its dictionary $D_j$ and it represents the learned frequency of each token $w_i \in D_j$. Both $D_j$ and $\vectorbold{f}_j$ are learned parameters of the LoFI layer.

The last key element of our approach is the computation of the histogram of the input node features in $\mathcal{N}_v$ with respect to the dictionary $D_j$, denoted as $ \vectorbold{h}_{ v,j } = h(X_v;D_j)$. 
 If the node features were discrete, the histogram would be easily obtained by counting the occurrences of node labels from the learned dictionary $D_j$.
Unfortunately, a discrete histogram is unsuitable for continuous optimization. Therefore, we relax the formulation by using a \textit{soft histogram}. 

For each node feature $\vectorbold{x}_u \in X_v$, we compute the normalized similarity score between $\vectorbold{x}_u$ and each $\mathbf{w}_i \in D_j $ by applying the softmax operation to the cosine similarities, \textit{i.e.},

\begin{equation}\label{eq:softmax} 
s(\vectorbold{x}_u, \mathbf{w}_i) = \frac{e^{t \langle \vectorbold{x}_u, \mathbf{w}_i \rangle}}{\sum_k e^{t \langle \vectorbold{x}_u, \mathbf{w}_k \rangle}}\;,
\end{equation}

\begin{figure}[t]
\centering
\includegraphics[width=1\textwidth]{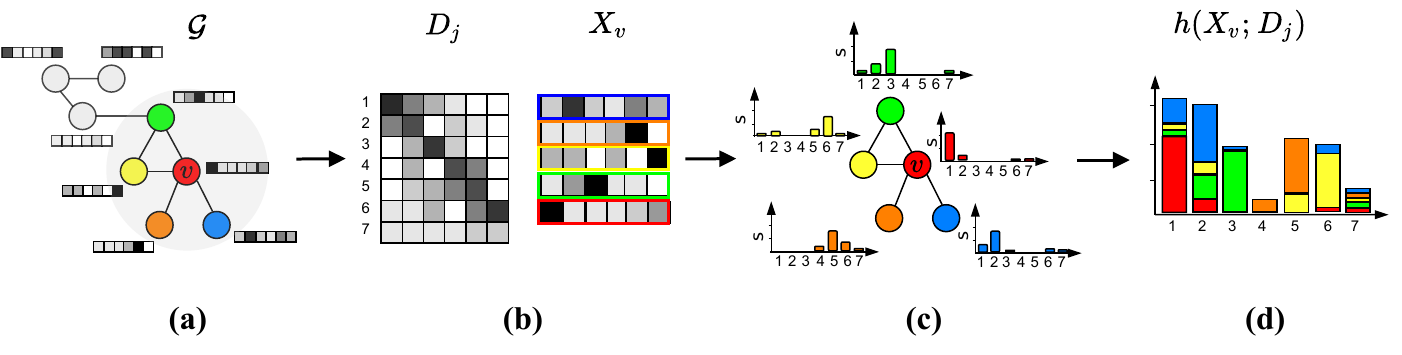}
\caption{Feature-based (soft) histogram computation, where we simplify the notation by omitting the layer number $l$. (a) Given a graph $\mathcal{G}$ with 8 nodes, we extract the 1-hop egonet $\mathcal{N}_v$, a subgraph of $\mathcal{G}$ centered on the vertex $v$, where its nodes are colour-coded for clarity. (b) $D_j$ is the $j$-th learned dictionary at the current layer. It has 7 entries (words) of size matching that of the node features.  $X_v$ represents the set of the 5 input node feature vectors associated to the 5 nodes of $\mathcal{N}_v$, colour-coded to match the corresponding nodes in (a). (c) For each node feature $\vectorbold{x}_u \in X_v$ we compute the normalized similarity score with respect to each dictionary token $\mathbf{w}_i \in D_j$. Here, we represent the similarity values for each feature vector $\vectorbold{x}_u$ by barplot. The height of a bar indicates the similarity of that feature vector $\vectorbold{x}_u$ to the word (words are $1,2,\dots,7$).
 (d) To obtain the feature (soft) histogram $h(X_v;D_j)$ of $X_v$, for each token $\mathbf{w}_i$ of the learned dictionary $D_j$, we sum the normalized similarities computed in (c) for every node feature $\vectorbold{x}_u \in X_v$.}
\label{fig:histogram_computation}
\end{figure}

where $t$ is a learnable temperature parameter. To interpret $\langle \cdot , \cdot \rangle$ as the cosine similarity we rescale both $\mathbf{x}_u$ and $\mathbf{w}_i$ to have unitary norm. Finally, the feature soft histogram representing $X_v$ is obtained by summing  the normalized similarity of all node features $\vectorbold{x}_u \in X_v$ for each $\mathbf{w}_i$, leading to a soft histogram whose elements are:

\begin{equation}\label{eq:histogram}    
   [ \vectorbold{h}_{ v,j } ]_i\;= \sum_{\vectorbold{x}_u \in X_v}{s(\vectorbold{x}_u, \mathbf{w}_i)} \;,
\end{equation}

where $[ \vectorbold{h}_{ v,j } ]_i$ is the $i$-th element of the vector $\vectorbold{h}_{ v,j }$. Note that, in the soft-histogram formulation, each $\mathbf{w}_i\in D_j$ is no longer required to be a discrete label, but it is a vector in the same space of the node features. Further, in the case of a set of orthonormal features (\textit{e.g.}, one-hot encoding) both in the dictionary and in the input graph, the soft-histogram formulation leads to the same representation as the discrete one.  Figure~\ref{fig:histogram_computation} shows the histogram computation procedure.

The intersection kernel, that relates to the tuple $(\vectorbold{h}_{ v,j}, D_j, \vectorbold{f}_j)$, 
is then computed in terms of the absolute difference between the two  histograms     $\vectorbold{h}_{ v,j }$ and $\vectorbold{f}_j$,  \textit{i.e.},

\begin{equation}\label{eq:mask_out}
K(\vectorbold{h}_{v,j} , \vectorbold{f}_j) = \frac{1}{2}\left( \|\vectorbold{h}_{v,j }\|_1\; + \;  \|\vectorbold{f}_j\|_1 \;-\; \|\vectorbold{h}_{v,j}- \vectorbold{f}_j\|_1\right)\,
\end{equation}
 
\begin{figure}
\centering
\includegraphics[width = 0.99\textwidth]{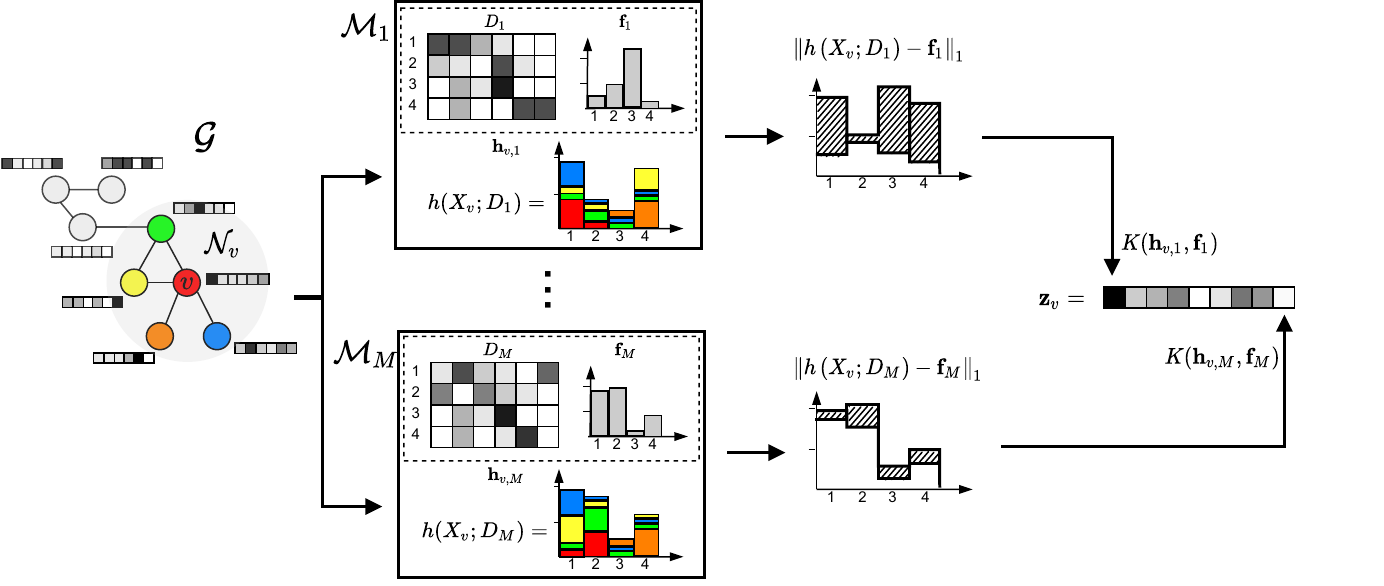}
\caption{The histogram intersection kernel at layer $l$, where we simplify the notation by omitting the layer number $l$. Given a node $v$ in $\mathcal{N}_v$ and the associated histograms $\vectorbold{h}_{v,j} = h(X_v;D_j)$, the  histogram intersection operation is repeated for each mask $\mathcal{M}_j = (D_j,\vectorbold{f}_j)$, where the $j$-th mask is the pair of learned histogram $\vectorbold{f}_j$ and dictionary $D_j$. The intersection between $h(X_v;D_j)$ and $\vectorbold{f}_j$ is defined in terms of the absolute difference between the two histograms. The obtained positive real value represents the similarity between the feature distribution associated to the node $v$ and the learned histogram, which in turn accounts for the $j$-th value of the updated node feature $\mathbf{z}_v$.}
\label{fig:histogram_intersection}
\end{figure}

With the soft histograms to hand, for each node $v$,  we can finally compute the updated node feature $\mathbf{z}_v \in \mathbb{R}^{+M}$ as
\begin{equation}
\mathbf{z}_v = [K(\vectorbold{h}_{ v,1 },\vectorbold{f}_1), \cdots , K(\vectorbold{h}_{ v,M },\vectorbold{f}_M)]\\.
\end{equation}
where $M$ is the total number of masks in a given layer.
The intersection operation along with the node feature construction is illustrated in Figure~\ref{fig:histogram_intersection}.  

Note that the computation of $\mathbf{z}_v$ is differentiable and thus we can learn the masks $\mathcal{M}_j = (D_j, \vectorbold{f}_j)$ for all layers through a standard backpropagation procedure.

\subsection{Relation with message passing GNNs}
In this subsection, we attempt to provide an interpretation of the proposed LoFI layer by viewing it through the lens of standard message passing graph neural networks. In particular, we show how our network, when the egonets radius $r=1$, can be interpreted as performing a version of message passing with non-linear aggregation.

To start, note that the message broadcasted by each node is the soft histogram of the node labels, \textit{i.e.}, the softmax of the cosine similarity between the node features $\vectorbold{x}_v$ and the {\em learned} dictionary entry $\mathbf{w}_i$. This can be seen as a linear transformation with a learned parameter followed by a non-linear transformation. The accumulation, on the other hand, is obtained by linearly accumulating the soft histograms from neighbouring nodes and then the non-linear computation of the intersection kernel against another learned parameter, namely the feature histogram $\vectorbold{f}_j$. 

More formally, let us replace the edges of the input graph with a new set of edges $\mathcal{E}'$ that connects the root $v$ of each egonet $\mathcal{N}_v$ to its neighbours, \textit{i.e.}, $(u,v) \in \mathcal{E}'$ if $u \in \mathcal{N}_v$, for each $v \in \mathcal{V}$. Recall the definition of the message passing mechanism in terms of the \textit{message} and \textit{update} functions in Eq.~\ref{eq:convolution}.

In our setting, the \textit{message} is the normalized similarity (Eq.~\ref{eq:softmax}), while the \textit{update} is given by the aggregation over the neighbours of the message (Eq.~\ref{eq:histogram}) followed by the non-linear kernel computation (Eq.~\ref{eq:mask_out}).

This in turn shows that our model can be interpreted as a type of message passing network. We posit however that the added non-linearities, together with the learned feature histogram, allow us to capture more complex and non-linearly separable properties of the neighbourhood.

\subsection{Computational complexity}
Let $n$ and $|\mathcal{E}|$ be the number of nodes and edges of $G$, respectively, $d$ the input features dimension, $M$ the embedding space dimension, \textit{i.e.}, the number of masks, and $W$ the number of words in each dictionary. Let $\overline{|\mathcal{N}_G|}$ denote the average size of the egonets computed over $G$, then the total number of nodes in all egonets is $n\overline{|\mathcal{N}_G|}$.

The complexity of a LoFI layer can be broken down into two components, corresponding to the computation of the normalized similarity (Eq.~\ref{eq:histogram}) and the histogram construction. The complexity of computing the normalized similarity is dominated by the dot product between each node feature and the dictionary words for all the $M$ masks, \textit{i.e.}, $O(ndMW)$. The histogram construction, on the other hand, requires summing, for each node and each mask, the normalized similarity (a vector of $W$ components) of all nodes belonging to the egonet. This in turn leads to a complexity of $O(n\overline{|\mathcal{N}_G|}MW)$. Finally, the computational complexity of the histogram kernel score $O(nMW)$ is negligible, giving a total complexity of $O(ndMW) + O(n\overline{|\mathcal{N}_G|}MW)$.

In contrast, the complexity of GCN, one of the simplest message passing models, is given by a message building complexity $O(ndM)$ and a aggregation complexity of $O(2|\mathcal{E}|M)$, where $M$ is the embeddings space dimension of a GCN layer. Our model has therefore the same $O(n)$ linear complexity wrt the number of nodes $n$, while the complexity wrt the number of edges depends on the egonet radius and the connectivity of the graph. For 1-hop egonets, where $n\overline{|\mathcal{N}_G|} = 2|\mathcal{E}|$, the complexity of GCN and our LoFI are both $O(|\mathcal{E}|)$.

\section{Experiments}\label{sec:experiments}
We run an extensive set of experiments to 1) measure the influence of the model parameters, 2) understand to what extent the learned histograms allow us to interpret the model decisions, 3) evaluate the performance of our model on both graph classification and graph regression datasets, and 4) study the time complexity of the proposed method. To ensure the reproducibility of our results, we make our code available on an online public repository\footnote{\url{https://github.com/gdl-unive/Histogram-Intersection-Kernel}}.

\subsection{Datasets and experimental setup}
We run our experiments on 7 widely used graph classification and regression datasets. For the graph classification task, we use 4 bio/chemo-informatics datasets collecting molecular graphs (MUTAG, NCI1, PROTEINS, and PTC)~\cite{kersting2016benchmark} and 2 datasets describing co-appearances of actors in movies (IMDB-B, IMDB-M)~\cite{yanardag2015deep}. Note that the 4 bio/chemo-informatics datasets all have categorical node features, while IMDB-B and IMDB-M have no node features. For the graph regression task, we use the drug-constrained solubility prediction dataset ZINC~\cite{irwin2012zinc}. We use a subset of the ZINC molecular graphs (12K out 250K), as in~\cite{dwivedi2020benchmarking}, to regress the desired molecular property, \textit{i.e.}, the constrained solubility. Each molecular graph has categorical node labels and although the dataset also includes information on the type of bond between nodes, we do not utilise it in our experiments.

We compare our model against: 1) eight state-of-the-art GNNs: 
GraphSAGE~\cite{hamilton2017inductive}, ECC~\cite{simonovsky2017dynamic}, DGCNN~\cite{zhang2018end}, DiffPool~\cite{ying2018hierarchical}, GIN~\cite{xu2018powerful}, (s)GIN~\cite{di2020mutual}, FSA-GNN~\cite{bicciato2022classifying}, and GCN~\cite{kipf2016semi}; 2) two distinct baselines, depending on the dataset type (see below): Molecular Fingerprint~\cite{ralaivola2005graph,luzhnica2019graph} and Deep Multisets~\cite{zaheer2017deep};  3) the WL kernel~\cite{shervashidze2011weisfeiler}; 4) RWGNN~\cite{nikolentzos2020random} and KerGNN~\cite{feng2022kergnns}, two GNN models employing a differentiable graph kernel. For the WL subtree kernel we use a $C$-SVM~\cite{chang2011libsvm} classifier. As baselines, we use the Molecular Fingerprint technique~\cite{ralaivola2005graph,luzhnica2019graph} for the 6 bio/chemo-informatics datasets and the permutation invariant model of~\cite{zaheer2017deep} for the IMDB datasets.

To ensure a fair comparison, we follow the same experimental protocol for each of these methods. For all datasets except ZINC, we perform 10-fold cross-validation where in each fold the training set is further subdivided into training and validation with a ratio of 9:1. The validation set is used for both early stopping and to select the best model within each fold. Importantly, folds and train/validation/test splits are consistent among all the methods. For the ZINC dataset, we use the train/validation/test split provided with the dataset (\textit{i.e.}, 10k train, 1k validation, 1k train). 
For all the methods, we perform a grid search to optimize the hyper-parameters. In particular, for the WL method, we optimize the value of $C$ and the number of WL iterations $h \in \{4, 5, 6, 7\}$. For all the comparative methods we investigate the hyper-parameter ranges proposed by the respective authors.
For our model, we explore the following hyper-parameters: number of layers in \{1, 3, 5, 7, 9, 11, 13\}, dropout in \{0, 0.1\}, egonet radius in \{1, 2, 3\}, number of masks \{4, 8, 16, 32\}, and dictionary size \{6, 8, 16, 32\}.

\begin{figure}[t!]
\centering
\begin{subfigure}{.5\textwidth}
\centering
\begin{overpic}[trim=18mm 0 24mm 11mm, clip, width=.9\textwidth]{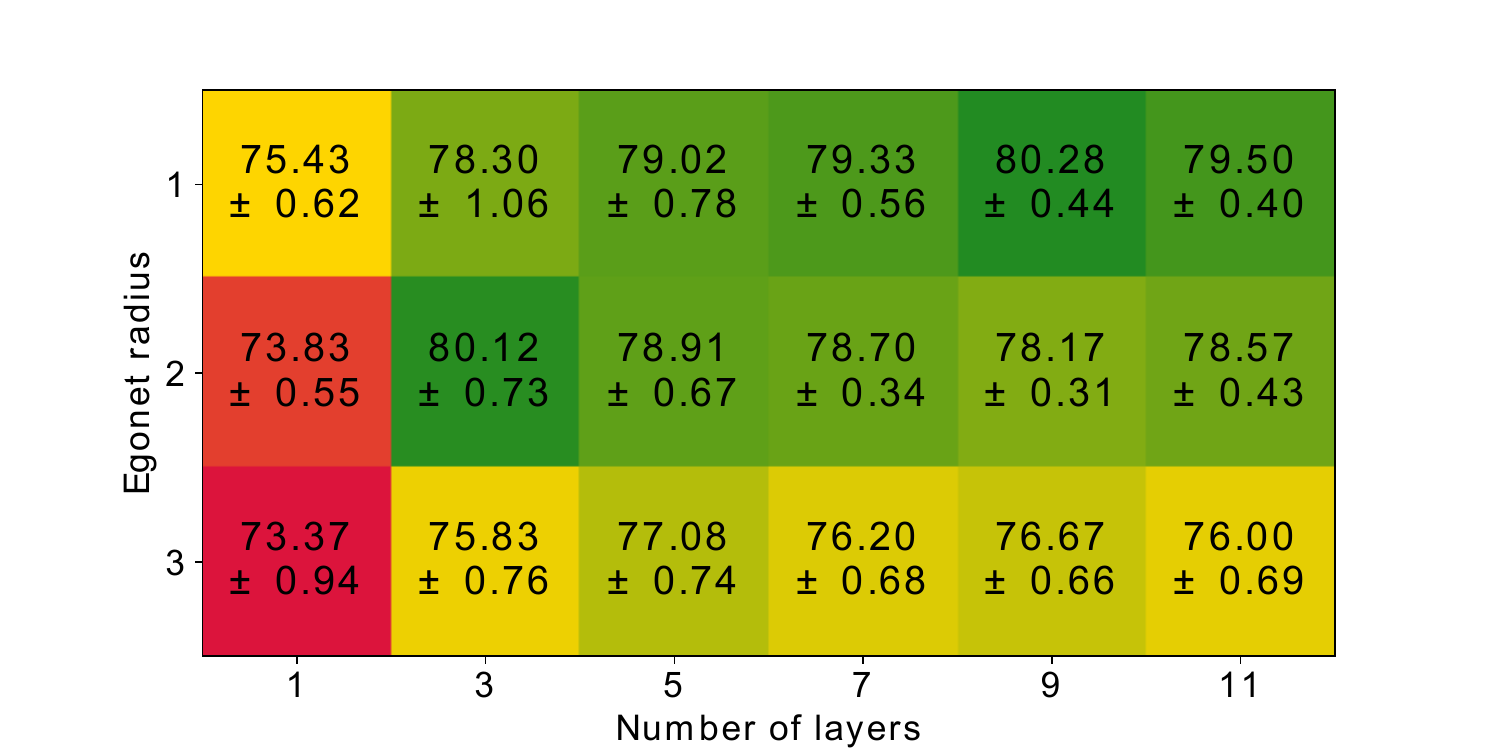}
\end{overpic}
\caption{NCI1}
\label{fig:ablation_layers_hops_NCI1}
\end{subfigure}%
\begin{subfigure}{.5\textwidth}
\centering
\begin{overpic}[trim=18mm 0 24mm 11mm, clip, width=.9\textwidth]{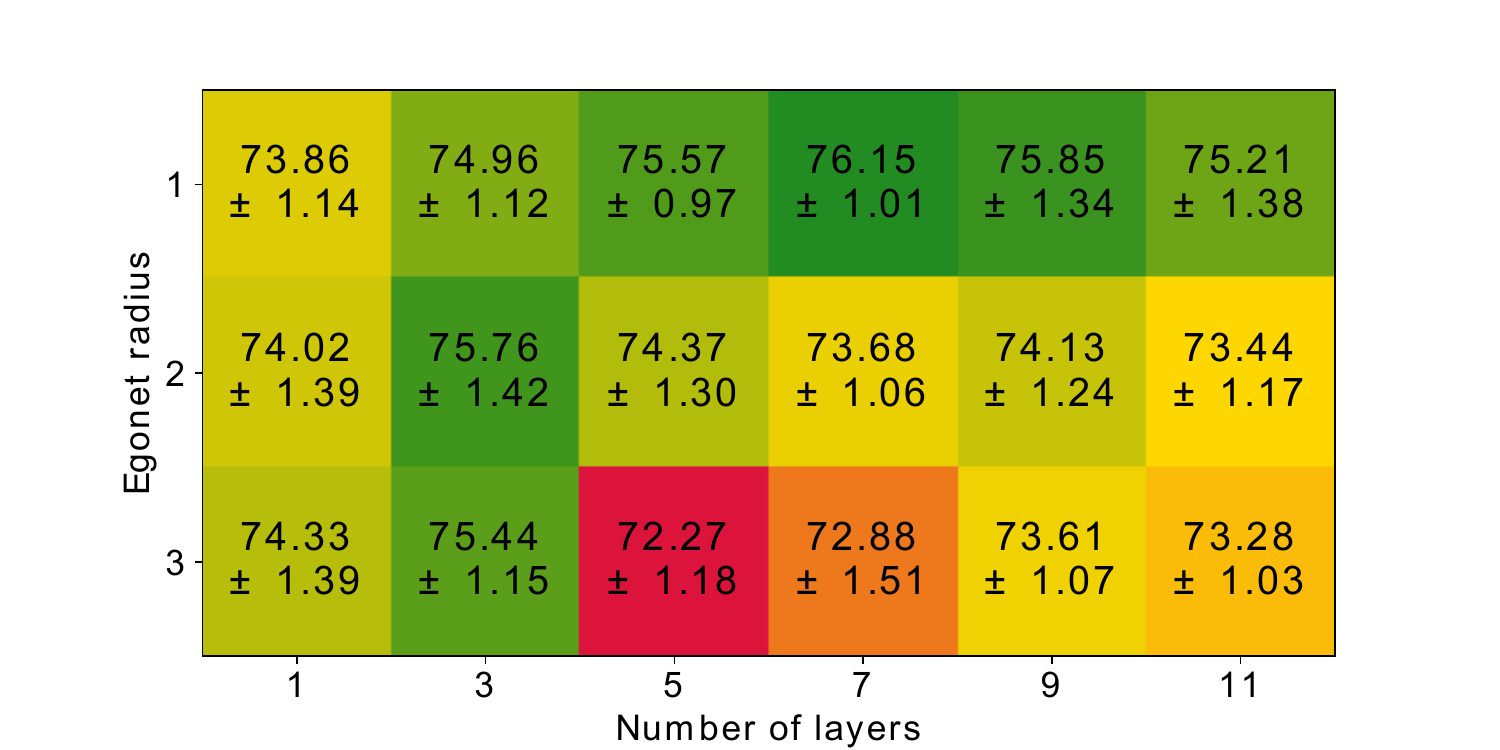}
\end{overpic}
\caption{PROTEINS}
\label{fig:ablation_layers_hops_proteins}
\end{subfigure}
\caption[short]{Average classification accuracy on (a) NCI1 and (b) PROTEINS dataset as we vary both the egonet radius and the number of layers.}
\label{fig:ablation_layers_hops}
\end{figure}

\begin{figure}[t!]
\centering
\begin{subfigure}{.5\textwidth}
\centering
\includegraphics[width=.8\textwidth]{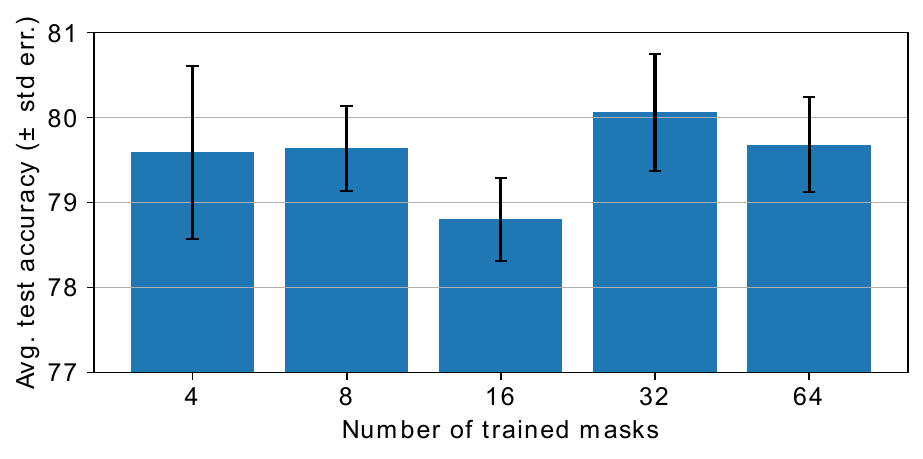}
\includegraphics[width=.8\textwidth]{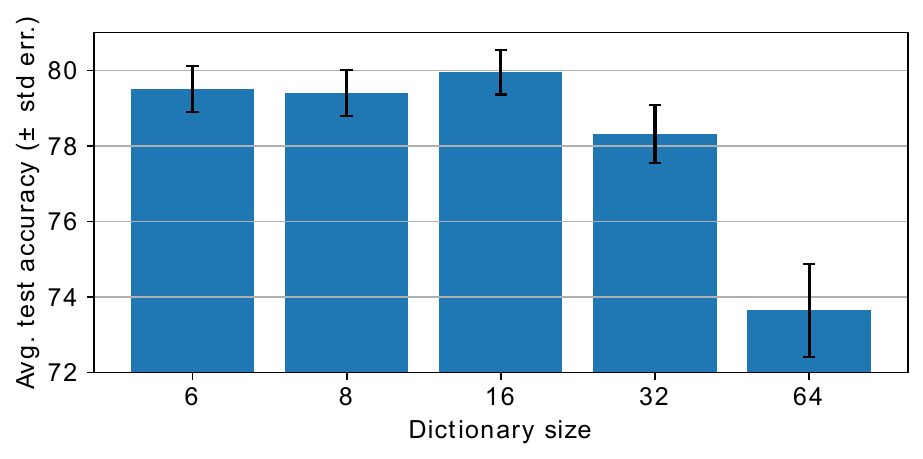}
\caption{NCI1}
\label{fig:ablation_learned_features_NCI1}
\end{subfigure}%
\begin{subfigure}{.5\textwidth}
\centering
\includegraphics[width=.8\textwidth]{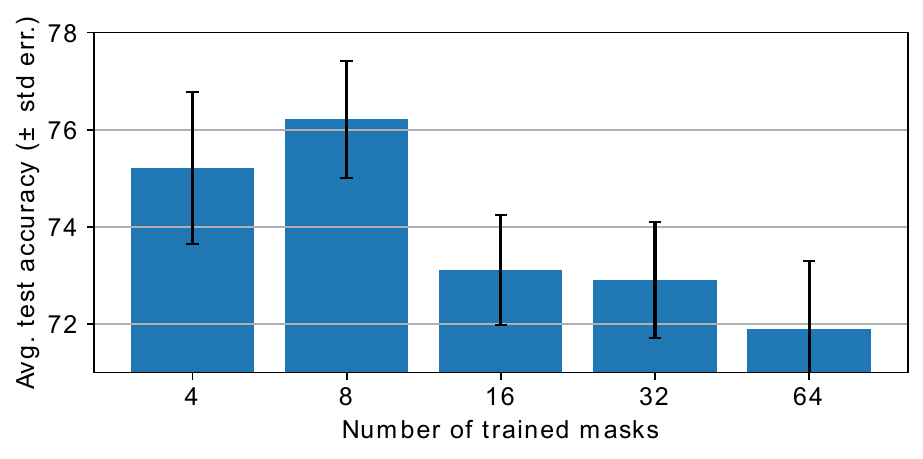}
\includegraphics[width=.8\textwidth]{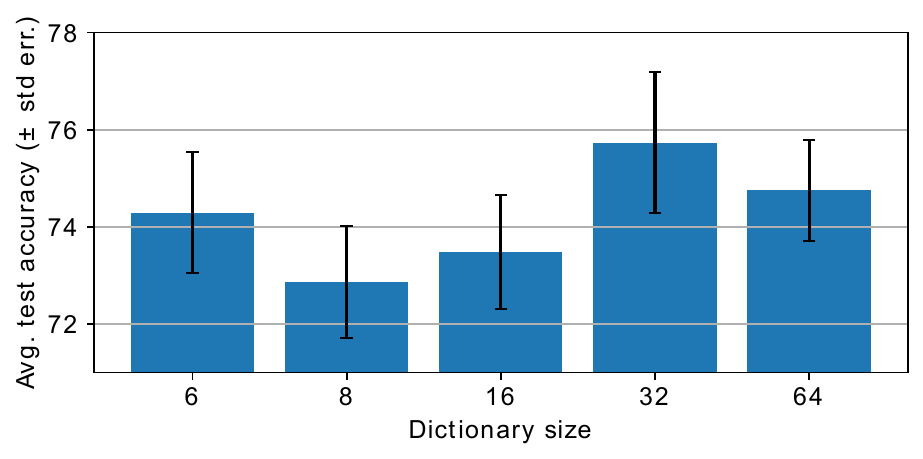}
\caption{PROTEINS}
\label{fig:ablation_learned_features_proteins}
\end{subfigure}
\caption[short]{Average classification accuracy on (a) NCI1 and (b) PROTEINS dataset comparing the number of trained masks (top) and the dictionary size (bottom).}
\label{fig:ablation_learned_features}
\end{figure}

In all our experiments, we train the model for $2000$ epochs, using the Adam optimizer with a learning rate of $0.001$ and a batch size of 32. The MLP takes as input the sum pooling of the node features and is composed of two layers of output dimension $m$ and $c$ (\# classes) with a ReLU activation in-between. We use cross-entropy and mean absolute error (MAE) losses for the graph classification and regression tasks, respectively.

\subsection{Ablation study}
We perform an ablation study of the network hyper-parameters on the PROTEINS and NCI1 datasets. To this end, we perform a grid search where we average the results over the 3 best performing models for each fold. Specifically, we study the influence of 1) the number of layers, 2) the egonet radius, 3) the number of trained masks, and 4) the dictionary size.

Figure~\ref{fig:ablation_layers_hops} shows how the average accuracy varies as a function of both the egonet radius and the number of layers, for both NCI1 (\ref{fig:ablation_layers_hops_NCI1}) and PROTEINS (\ref{fig:ablation_layers_hops_proteins}). We observe that a network with a smaller egonet radius requires more depth to propagate the feature information far enough on the graph, while as the egonet radius increases, the optimal number of layers tends to be lower.

Figure~\ref{fig:ablation_learned_features} shows instead how the classification accuracy varies on NCI (\ref{fig:ablation_learned_features_NCI1}) and PROTEINS (\ref{fig:ablation_learned_features_proteins}) as we change the number of masks (top) and the dictionary size (bottom). We do not observe a clear trend in this case, with the extent of the influence of these hyper-parameters on the network performance depending on the dataset.

\begin{figure}[t!]
\centering
\includegraphics[width=0.99\textwidth]{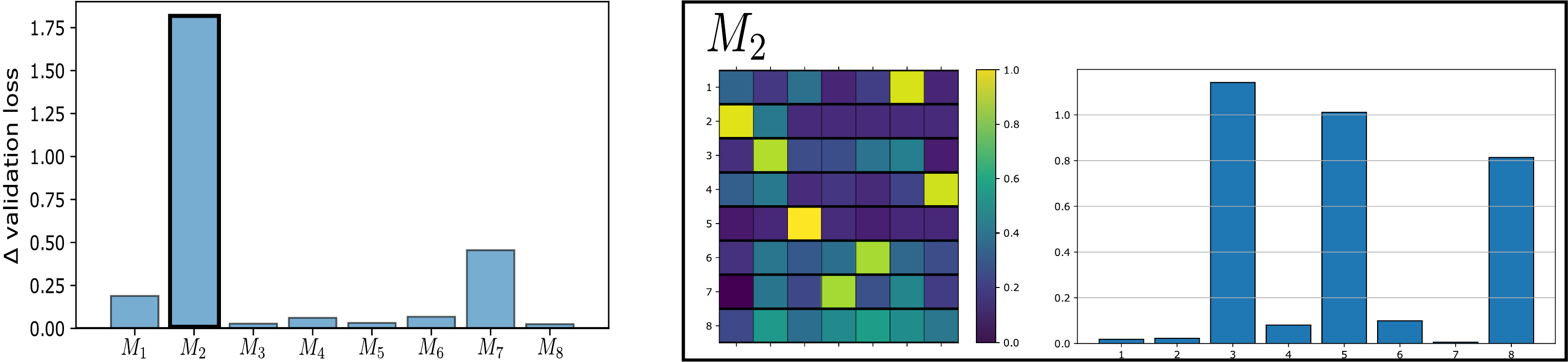}
\vspace{0.1in}
\caption{Learned masks interpretation. For this example, we trained a model with 8 masks on the MUTAG dataset (7 discrete node labels). Only the first layer of the model is considered, where features are one-hot encoded vectors. Left: the influence of each mask on the validation loss. The histogram shows how mask $\mathcal{M}_2$  is the most impacting one.  Right: the dictionary $D_2$ (left) and the learned histogram $\vectorbold{f}_2$ (right) associated with $\mathcal{M}_2$. The learned histogram $\vectorbold{f}_2$  supports a neighbourhood whose label composition is mostly of labels 2 and 3 (dictionary words $3$ and $5$), roughly in equal quantities.}
\label{fig:interpret}
\end{figure}

\subsection{Learned masks visualization}
The output of our convolution operator is a similarity score between the local distribution of node features and the learned histograms. This allows one to interpret  the output of our model, in contrast to the standard message passing methods that usually involve non-trivial non-linear transformations and aggregation of the graph features. 

In Figure~\ref{fig:interpret} we show the most significant mask from the first layer of a model trained on MUTAG, where the number of learned masks was set to 8. To identify the most significant mask, we iteratively disable one mask by setting its output (\textit{i.e.}, the result of Eq.~\ref{eq:mask_out}) to zero. This in turn allows us to measure the impact on the validation loss (histogram on the left). We can clearly see that the second mask $\mathcal{M}_2$ is the most significant one, as it has the highest impact on the validation performance. On the right-hand side of Figure~\ref{fig:interpret} we show the dictionary $D_2$ and learned histogram $\vectorbold{f}_2$ associated with $\mathcal{M}_2$. The results suggest that the learned dictionary tends to learn feature vectors similar to the input ones. For example, in the first layer, the features are one-hot encoded vectors of the 7 node labels in the dataset, which results in a learned dictionary with the first 7 entries (words) peaked on the individual feature dimensions. Interestingly, the eighth dictionary word is almost uniform over the node features. As we can see from the learned histogram, the model is learning a graph-neighbourhood with almost exclusively dictionary words 3, 5, and 8. Word 8, corresponding to a uniform distribution over the node features, gives a result in its histogram entry proportional to the size of the egonet, or the degree of the node under analysis. On the other hand, dictionary words 3 and 5 correspond to node labels 2 and 3. This means that the learned histogram $\vectorbold{f}_2$ gives the maximum result to a neighbourhood of a particular size (corresponding to the height of histogram entry 8), whose label composition is mostly of label 2 (corresponding to dictionary word 3) and label 3 (\textit{i.e.}, word 5), roughly in equal quantities.

\begin{table*}[t!]
\caption{Classification (mean accuracy $\pm$ standard error) and regression (MAE) results on the 7 datasets considered in this study. The best performance (per dataset) is highlighted in bold, and the second best is underlined. We indicate with $C$ and $R$ the classification and regressions tasks, respectively.}
\label{tab:comparisons}
\centering
{\fontsize{7.2pt}{7.2pt}\fontfamily{ppl}\selectfont
\begin{tabular*}{\textwidth}{l|cccccc|c}
\toprule
g& MUTAG(C) & PTC(C) & NCI1(C) & PROTEINS(C) & IMDB-B(C) & IMDB-M(C) & ZINC(R) \\ \hline
Baseline & 78.57 ± 4.00 & 58.34 ± 2.02 & 68.50 ± 0.87 & 73.05 ± 0.90 & 49.50 ± 0.79 & 32.00 ± 1.17 & 0.66 \\
WL & 82.67 ± 2.22 & 55.39 ± 1.27 & \underline{79.32 ± 1.48} & 74.16 ± 0.38 & \underline{71.80 ± 1.03}  & 43.33 ± 0.56  & 1.24 \\ \hline
GCN & 83.95 ± 3.33 & 55.34 ± 2.22 & 62.51 ± 0.63 & 70.63 ± 1.52 & 49.30 ± 0.45 & 33.27 ± 0.24 & 0.51\\
DiffPool & 81.35 ± 1.86 & 55.87 ± 2.73 & 75.72 ± 0.79 & 73.13 ± 1.49 & 67.80 ± 1.44 & 44.93 ± 1.02 & 0.49 \\
GIN & 78.13 ± 2.88 & 56.72 ± 2.66 & 78.63 ± 0.82 & 70.98 ± 1.61 & 71.10 ± 1.65 & 46.93 ± 1.06 & 0.38 \\
DGCNN & \underline{85.06 ± 2.50} & 53.50 ± 2.71 & 76.56 ± 0.93 & 74.31 ± 1.03 & 53.00  ± 1.32 & 41.80 ± 1.39 & 0.60 \\
ECC & 79.68 ± 3.78 & 54.43 ± 2.74 & 73.48 ± 0.71 & 73.76 ± 1.60& 68.30 ± 1.56 & 44.33 ± 1.59 & 0.73 \\
GraphSAGE & 77.57 ± 4.22 & 59.87 ± 1.91 & 75.89 ± 0.96 & 73.11 ± 1.27 & 68.80  ± 2.26 & 47.20 ± 1.18 & 0.46 \\
sGIN & 84.09 ± 1.72 & 56.37 ± 2.28 & 77.54 ± 1.00 & 73.59 ± 1.47 &  71.30 ± 1.75 &
\underline{48.60 ± 1.48} & \underline{0.37} \\
SFA-GNN & 84.04 ± 2.48 & \underline{63.18} ± 3.63 & 78.51 ± 0.40 & \textbf{75.13 ± 1.53} & 62.42 ± 2.20 & 35.95 ± 1.60 & \underline{0.37} \\
\hline 
KerGNN & 82.43 ± 2.73 & 53.15 ± 1.83 & 74.16 ± 3.36 & 72.96 ± 1.46 & 67.90 ± 1.33 & 44.87 ± 1.52 & 0.57 \\
RWGNN & 82.51 ± 2.47 & 55.47 ± 2.70 & 72.94 ± 1.16 & 73.95 ± 1.32 & 69.90 ± 1.32 & 46.20 ± 1.08 & 0.55 \\ \hline
\textbf{GNN-LoFI} & \textbf{86.78 ± 2.00} & \textbf{65.73 ± 5.29} & \textbf{79.45 ± 0.47} & \underline{75.10 ± 1.04} & \textbf{76.56 ± 2.12} & \textbf{48.80 ± 1.23} & \textbf{0.27} \\
\hline
\end{tabular*}}
\end{table*}


\subsection{Graph classification and regression results}
Table~\ref{tab:comparisons} shows the results of the experimental evaluation on real-world datasets. We outperform all competing approaches in all the graph classification datasets considered in this study. We also outperform all competing approaches on the ZINC dataset. Interestingly, we achieve very high classification accuracy on both IMDB-B and IMDB-M, when compared to the other methods. This is despite the fact that GNN-LoFI relies heavily on the node labels, which are not available on these two datasets. Instead, we assign the same initial label (\textit{i.e.}, $0$) to all nodes of the input graphs and we let our network compute new labels with the procedure described in Section~\ref{sec:model}. The excellent performance obtained on the IMDB datasets, therefore, suggests that GNN-LoFI is able to successfully capture the structural information of the graph when the node feature information is missing. 

\begin{table}[t]
\centering
\caption{Average time (± standard deviation) required at training (per epoch) and inference (per sample) by our method, for some of the datasets considered in this study. \chg{The GNN-GIN and GNN-GCN columns report the training time and inference time of comparable models where our convolution operation has been replaced by GIN and GCN convolutions, respectively}. The two rightmost columns show some relevant datasets statistics that influence the running time.}\label{tab:speed}
\fontsize{7.2pt}{7.2pt}\fontfamily{ppl}\selectfont
\setlength{\tabcolsep}{4.1pt}
\begin{tabular}{l | >
{\centering\arraybackslash}m{0.1\linewidth} >{\centering\arraybackslash}m{0.1\linewidth} | >{\centering\arraybackslash}m{0.1\linewidth} >{\centering\arraybackslash}m{0.1\linewidth}  | >{\centering\arraybackslash}m{0.1\linewidth} >{\centering\arraybackslash}m{0.1\linewidth}| >{\centering\arraybackslash}m{0.05\linewidth} >{\centering\arraybackslash}m{0.05\linewidth}}
\toprule
\multicolumn{1}{c}{ }&\multicolumn{2}{c}{GNN-LoFI} & \multicolumn{2}{c}{GNN-GIN} & \multicolumn{2}{c}{\chg{GNN-GCN}}  & \\
& Training time per epoch (s)& Inference time per sample (ms) & Training time per epoch (s)& Inference time per sample (ms) & \chg{Training time per epoch (s)}& \chg{Inference time per sample (ms)} & \# Graphs & Avg \# Nodes \\\midrule
PROTEINS &  4.10 ± 1.62  & 0.86 ± 0.06 & 4.03 ± 1.41 & 0.83 ± 0.08 & \chg{2.20 ± 0.25} & \chg{1.00 ± 0.07}&1113 &39.06 \\
MUTAG    &  0.55 ± 0.17  &0.50 ± 0.09 & 0.73 ± 0.44  & 0.47 ± 0.15 & \chg{0.49 ± 0.06} & \chg{0.53 ± 0.08}&188 &17.93 \\
PTC &  0.94 ± 0.33  &0.53 ± 0.09 & 0.89 ± 0.14 & 0.49 ± 0.04 & \chg{0.83 ± 0.05} & \chg{0.74 ± 0.05}&344 &14.69 \\
NCI1 &  12.52 ± 3.60  &0.67 ± 0.06 & 15.95 ± 7.76 & 0.80 ± 0.17 & \chg{8.16 ± 0.99} & \chg{0.80 ± 0.09}&4110 &29.87 \\
IMDB-B &  4.98 ± 0.90  & 0.48 ± 0.05 & 3.99 ± 1.12 & 0.69 ± 0.07 & \chg{2.21 ± 0.25} & \chg{0.60 ± 0.05}& 1000 &19.77 \\
\bottomrule
\end{tabular}
\end{table}

\subsection{Runtime evaluation}
We report in Table~\ref{tab:speed} the average training time per dataset (per epoch) of the best models for each fold which were selected during the grid search. Both training and testing of each model were performed using a TitanXp with 12 GiB of memory. \chg{To better relate the time complexity of our method to that of alternative ones, we compare our architecture with similar ones where our convolution is replaced by the GIN and GCN convolutions, respectively}. We keep the same number of layers as our models and use an MLP with a comparable number of parameters as our LoFI layer. We can see that the time complexity of our method is comparable with that of classical GNNs.

\section{Conclusion}\label{sec:conclusion}
In this paper, we proposed a novel graph neural network where the standard message passing mechanism is replaced by an analysis of the local distribution of node features. This is achieved by extracting the distribution of the node features in egonets centered around each node of the graph, which is then compared to a set of learned label distributions via a differentiable soft-histogram intersection. For each node, the resulting similarity information is propagated to its neighbours, which in turn can be interpreted as a message passing mechanism where the message is determined by the ensemble of the features. We validated our model through an extensive set of experiments, showing that it can outperform widely used alternative methods. One of the limitations of our method is that, although it captures the local topology of the graph by looking at feature distributions over neighbourhoods, it does not directly encapsulate the graph structure and thus the learned masks do not contain information on the presence of potentially discriminative structural patterns. Future work will address this issue by allowing our model to jointly learn the distribution of features as well as the structure. Our model is specifically designed for graph-level rather than node-level classification tasks. The latter are usually performed on graphs, such as social networks, where structural information plays a secondary role wrt to node features, which in turn are typically continuous-valued and high dimensional. Specifically, the use of histograms in our model introduces boundary effects due to the inability to capture the correlation between adjacent discretization buckets, especially in high dimensional spaces. We plan to address this shortcoming in future work by employing for example the Earth mover’s distance.

\section{Acknowledgments}

This project is partially supported by the PRIN 2022 project n. 2022AL45R2 (EYE-FI.AI, CUP H53D2300350-0001). G.M. acknowledges financial support from project iNEST (Interconnected NordEst Innovation Ecosystem), funded by European Union Next - GenerationEU - National Recovery and Resilience Plan (\textit{NRRP}) - \texttt{MISSION 4 COMPONENT 2, INVESTMENT N. ECS00000043 - CUP N. H43C22000540006}. In this regard, the manuscript reflects only the authors’ views and opinions, neither the European Union nor the European Commission can be considered responsible for them.





\bibliographystyle{elsarticle-num} 
\bibliography{biblio}

\begin{thebibliography}{10}
\expandafter\ifx\csname url\endcsname\relax
  \def\url#1{\texttt{#1}}\fi
\expandafter\ifx\csname urlprefix\endcsname\relax\def\urlprefix{URL }\fi
\expandafter\ifx\csname href\endcsname\relax
  \def\href#1#2{#2} \def\path#1{#1}\fi

\bibitem{hochreiter1997long}
S.~Hochreiter, J.~Schmidhuber, Long short-term memory, Neural computation 9~(8)
  (1997) 1735--1780.

\bibitem{hinton2007learning}
G.~E. Hinton, Learning multiple layers of representation, Trends in cognitive
  sciences 11~(10) (2007) 428--434.

\bibitem{bengio2012practical}
Y.~Bengio, Practical recommendations for gradient-based training of deep
  architectures, in: Neural networks: Tricks of the trade, Springer, 2012, pp.
  437--478.

\bibitem{oh2004gpu}
K.-S. Oh, K.~Jung, Gpu implementation of neural networks, Pattern Recognition
  37~(6) (2004) 1311--1314.

\bibitem{vaswani2017attention}
A.~Vaswani, N.~Shazeer, N.~Parmar, J.~Uszkoreit, L.~Jones, A.~N. Gomez,
  {\L}.~Kaiser, I.~Polosukhin, Attention is all you need, Advances in neural
  information processing systems 30 (2017).

\bibitem{devlin2019BERT}
J.~Devlin, M.~Chang, K.~Lee, K.~Toutanova,
  \href{https://doi.org/10.18653/v1/n19-1423}{{BERT:} pre-training of deep
  bidirectional transformers for language understanding}, in: J.~Burstein,
  C.~Doran, T.~Solorio (Eds.), Proceedings of the 2019 Conference of the North
  American Chapter of the Association for Computational Linguistics: Human
  Language Technologies, {NAACL-HLT} 2019, Minneapolis, MN, USA, June 2-7,
  2019, Volume 1 (Long and Short Papers), Association for Computational
  Linguistics, 2019, pp. 4171--4186.
\newblock \href {https://doi.org/10.18653/v1/n19-1423}
  {\path{doi:10.18653/v1/n19-1423}}.
\newline\urlprefix\url{https://doi.org/10.18653/v1/n19-1423}

\bibitem{guo2022cmt}
J.~Guo, K.~Han, H.~Wu, Y.~Tang, X.~Chen, Y.~Wang, C.~Xu, Cmt: Convolutional
  neural networks meet vision transformers, in: Proceedings of the IEEE/CVF
  Conference on Computer Vision and Pattern Recognition, 2022, pp.
  12175--12185.

\bibitem{conwell2022testing}
C.~Conwell, T.~Ullman, Testing relational understanding in text-guided image
  generation, arXiv preprint arXiv:2208.00005 (2022).

\bibitem{johnson2015image}
J.~Johnson, R.~Krishna, M.~Stark, L.-J. Li, D.~Shamma, M.~Bernstein,
  L.~Fei-Fei, Image retrieval using scene graphs, in: Proceedings of the IEEE
  conference on computer vision and pattern recognition, 2015, pp. 3668--3678.

\bibitem{borgwardt2005protein}
K.~M. Borgwardt, C.~S. Ong, S.~Sch{\"o}nauer, S.~Vishwanathan, A.~J. Smola,
  H.-P. Kriegel, Protein function prediction via graph kernels, Bioinformatics
  21~(suppl\_1) (2005) i47--i56.

\bibitem{lima2014coding}
A.~Lima, L.~Rossi, M.~Musolesi, Coding together at scale: Github as a
  collaborative social network, in: Eighth international AAAI conference on
  weblogs and social media, 2014.

\bibitem{scarselli2008graph}
F.~Scarselli, M.~Gori, A.~C. Tsoi, M.~Hagenbuchner, G.~Monfardini, The graph
  neural network model, IEEE transactions on neural networks 20~(1) (2008)
  61--80.

\bibitem{atwood2016diffusion}
J.~Atwood, D.~Towsley, Diffusion-convolutional neural networks, in: Advances in
  neural information processing systems, 2016, pp. 1993--2001.

\bibitem{kipf2016semi}
T.~N. Kipf, M.~Welling, Semi-supervised classification with graph convolutional
  networks, in: Proceedings of the 5th International Conference on Learning
  Representations, ICLR '17, 2017.

\bibitem{gilmer2017neural}
J.~Gilmer, S.~S. Schoenholz, P.~F. Riley, O.~Vinyals, G.~E. Dahl, Neural
  message passing for quantum chemistry, in: International conference on
  machine learning, PMLR, 2017, pp. 1263--1272.

\bibitem{velickovic2018graph}
P.~Veli{\v{c}}kovi{\'{c}}, G.~Cucurull, A.~Casanova, A.~Romero, P.~Li{\`{o}},
  Y.~Bengio, Graph attention networks, International Conference on Learning
  Representations (2018).

\bibitem{bai2020learning}
L.~Bai, L.~Cui, Y.~Jiao, L.~Rossi, E.~Hancock, Learning backtrackless
  aligned-spatial graph convolutional networks for graph classification, IEEE
  Transactions on Pattern Analysis and Machine Intelligence (2020).

\bibitem{cosmo2021graph}
L.~Cosmo, G.~Minello, M.~Bronstein, E.~Rodol{\`a}, L.~Rossi, A.~Torsello, Graph
  kernel neural networks, arXiv preprint arXiv:2112.07436 (2021).

\bibitem{cosmo2020latent}
L.~Cosmo, A.~Kazi, S.-A. Ahmadi, N.~Navab, M.~Bronstein, Latent-graph learning
  for disease prediction, in: International Conference on Medical Image
  Computing and Computer-Assisted Intervention, Springer, 2020, pp. 643--653.

\bibitem{kazi2022differentiable}
A.~Kazi, L.~Cosmo, S.-A. Ahmadi, N.~Navab, M.~Bronstein, Differentiable graph
  module (dgm) for graph convolutional networks, IEEE Transactions on Pattern
  Analysis and Machine Intelligence (2022).

\bibitem{bicciato2022classifying}
A.~Bicciato, L.~Cosmo, G.~Minello, L.~Rossi, A.~Torsello, Classifying me
  softly: A novel graph neural network based on features soft-alignment, in:
  Joint IAPR International Workshops on Statistical Techniques in Pattern
  Recognition (SPR) and Structural and Syntactic Pattern Recognition (SSPR),
  Springer, 2022, pp. 43--53.

\bibitem{shervashidze2011weisfeiler}
N.~Shervashidze, P.~Schweitzer, E.~J. Van~Leeuwen, K.~Mehlhorn, K.~M.
  Borgwardt, Weisfeiler-lehman graph kernels., Journal of Machine Learning
  Research 12~(9) (2011).

\bibitem{xu2018powerful}
K.~Xu, W.~Hu, J.~Leskovec, S.~Jegelka, How powerful are graph neural networks?,
  arXiv preprint arXiv:1810.00826 (2018).

\bibitem{morris2019weisfeiler}
C.~Morris, M.~Ritzert, M.~Fey, W.~L. Hamilton, J.~E. Lenssen, G.~Rattan,
  M.~Grohe, Weisfeiler and leman go neural: Higher-order graph neural networks,
  in: Proceedings of the AAAI Conference on Artificial Intelligence, 2019, pp.
  4602--4609.

\bibitem{kriege2020survey}
N.~M. Kriege, F.~D. Johansson, C.~Morris, A survey on graph kernels, Applied
  Network Science 5~(1) (2020) 1--42.

\bibitem{zhang2018retgk}
Z.~Zhang, M.~Wang, Y.~Xiang, Y.~Huang, A.~Nehorai, Retgk: Graph kernels based
  on return probabilities of random walks, Advances in Neural Information
  Processing Systems 31 (2018).

\bibitem{borgwardt2005shortest}
K.~M. Borgwardt, H.-P. Kriegel, Shortest-path kernels on graphs, in: Fifth IEEE
  international conference on data mining (ICDM'05), IEEE, 2005, pp. 8--pp.

\bibitem{bevilacqua2022equivariant}
B.~Bevilacqua, F.~Frasca, D.~Lim, B.~Srinivasan, C.~Cai, G.~Balamurugan, M.~M.
  Bronstein, H.~Maron, Equivariant subgraph aggregation networks, in:
  International Conference on Learning Representations, 2022.

\bibitem{bouritsas2022improving}
G.~Bouritsas, F.~Frasca, S.~P. Zafeiriou, M.~Bronstein, Improving graph neural
  network expressivity via subgraph isomorphism counting, IEEE Transactions on
  Pattern Analysis and Machine Intelligence (2022).

\bibitem{nikolentzos2020random}
G.~Nikolentzos, M.~Vazirgiannis, Random walk graph neural networks, Advances in
  Neural Information Processing Systems 33 (2020) 16211--16222.

\bibitem{feng2022kergnns}
A.~Feng, C.~You, S.~Wang, L.~Tassiulas, Kergnns: Interpretable graph neural
  networks with graph kernels, in: Proceedings of the AAAI Conference on
  Artificial Intelligence, 2022.

\bibitem{muller2023graph}
E.~M{\"u}ller, Graph clustering with graph neural networks, Journal of Machine
  Learning Research 24 (2023) 1--21.

\bibitem{zhang2022non}
H.~Zhang, J.~Shi, R.~Zhang, X.~Li, Non-graph data clustering via
  $\mathcal{O}(n)$ bipartite graph convolution, IEEE Transactions on Pattern
  Analysis and Machine Intelligence (2022).

\bibitem{wu2020comprehensive}
Z.~Wu, S.~Pan, F.~Chen, G.~Long, C.~Zhang, S.~Y. Philip, A comprehensive survey
  on graph neural networks, IEEE transactions on neural networks and learning
  systems 32~(1) (2020) 4--24.

\bibitem{kersting2016benchmark}
K.~Kersting, N.~M. Kriege, C.~Morris, P.~Mutzel, M.~Neumann, Benchmark data
  sets for graph kernels, URL http://graphkernels. cs. tu-dortmund. de (2016).

\bibitem{yanardag2015deep}
P.~Yanardag, S.~Vishwanathan, Deep graph kernels, in: Proceedings of the 21th
  ACM SIGKDD international conference on knowledge discovery and data mining,
  2015, pp. 1365--1374.

\bibitem{irwin2012zinc}
J.~J. Irwin, T.~Sterling, M.~M. Mysinger, E.~S. Bolstad, R.~G. Coleman, Zinc: a
  free tool to discover chemistry for biology, Journal of chemical information
  and modeling 52~(7) (2012) 1757--1768.

\bibitem{dwivedi2020benchmarking}
V.~P. Dwivedi, C.~K. Joshi, T.~Laurent, Y.~Bengio, X.~Bresson, Benchmarking
  graph neural networks, arXiv preprint arXiv:2003.00982 (2020).

\bibitem{hamilton2017inductive}
W.~Hamilton, Z.~Ying, J.~Leskovec, Inductive representation learning on large
  graphs, Advances in neural information processing systems 30 (2017).

\bibitem{simonovsky2017dynamic}
M.~Simonovsky, N.~Komodakis, Dynamic edge-conditioned filters in convolutional
  neural networks on graphs, in: Proceedings of the IEEE conference on computer
  vision and pattern recognition, 2017, pp. 3693--3702.

\bibitem{zhang2018end}
M.~Zhang, Z.~Cui, M.~Neumann, Y.~Chen, An end-to-end deep learning architecture
  for graph classification, in: Thirty-Second AAAI Conference on Artificial
  Intelligence, 2018.

\bibitem{ying2018hierarchical}
R.~Ying, J.~You, C.~Morris, X.~Ren, W.~L. Hamilton, J.~Leskovec, Hierarchical
  graph representation learning with differentiable pooling, arXiv preprint
  arXiv:1806.08804 (2018).

\bibitem{di2020mutual}
X.~Di, P.~Yu, R.~Bu, M.~Sun, Mutual information maximization in graph neural
  networks, in: 2020 International Joint Conference on Neural Networks (IJCNN),
  IEEE, 2020, pp. 1--7.

\bibitem{ralaivola2005graph}
L.~Ralaivola, S.~J. Swamidass, H.~Saigo, P.~Baldi, Graph kernels for chemical
  informatics, Neural networks 18~(8) (2005) 1093--1110.

\bibitem{luzhnica2019graph}
E.~Luzhnica, B.~Day, P.~Li{\`o}, On graph classification networks, datasets and
  baselines, arXiv preprint arXiv:1905.04682 (2019).

\bibitem{zaheer2017deep}
M.~Zaheer, S.~Kottur, S.~Ravanbhakhsh, B.~P{\'o}czos, R.~Salakhutdinov, A.~J.
  Smola, Deep sets, in: Proceedings of the 31st International Conference on
  Neural Information Processing Systems, 2017, pp. 3394--3404.

\bibitem{chang2011libsvm}
C.-C. Chang, C.-J. Lin, Libsvm: a library for support vector machines, ACM
  transactions on intelligent systems and technology (TIST) 2~(3) (2011) 1--27.

\end{thebibliography}
\end{document}